\documentclass{article}

\usepackage{microtype}
\usepackage{graphicx}
\usepackage{subcaption}
\usepackage{booktabs} 

\usepackage{hyperref}

\usepackage{silence}
\usepackage[accepted]{icml2026}

\usepackage{amsmath}
\usepackage{amssymb}
\usepackage{mathtools}
\usepackage{amsthm}

\usepackage[capitalize,noabbrev]{cleveref}

\theoremstyle{plain}

\theoremstyle{definition}

\theoremstyle{remark}

\usepackage[textsize=tiny]{todonotes}

\usepackage{lipsum}
\usepackage{multirow}
\usepackage[table]{xcolor}
\usepackage{xspace}
\usepackage{algorithm}
\usepackage[normalem]{ulem}
\usepackage{silence}
\makeatletter
\DeclareRobustCommand\onedot{\futurelet\@let@token\@onedot}
\def\@onedot{\ifx\@let@token.\else.\null\fi\xspace}

\def\eg{\emph{e.g}\onedot} 
\def\ie{\emph{i.e}\onedot}

\makeatother

\definecolor{CoolGray1C}{HTML}{d9d8d6}
\definecolor{7621C}{HTML}{b12028}
\definecolor{7734C}{HTML}{266041}
\definecolor{280C}{HTML}{002169}

\definecolor{firstColor}{RGB}{165, 0, 52}     
\definecolor{secondColor}{RGB}{0, 87, 118}   

\newcommand{\first}[1]{\textbf{\textcolor{firstColor}{#1}}}
\newcommand{\second}[1]{\textcolor{secondColor}{\uline{#1}}}

\newcommand{\SA}{\textcolor{280C}{\textbf{\textit{Semantic Anchoring}}}\xspace}
\newcommand{\PI}{\textcolor{7734C}{\textbf{\textit{Primitive Imbuing}}}\xspace}
\newcommand{\CS}{\textcolor{7621C}{\textbf{\textit{Conceptual Steering}}}\xspace}

\newcommand{\sa}{\textcolor{280C}{\textbf{\textit{SA}}}\xspace}
\newcommand{\pI}{\textcolor{7734C}{\textbf{\textit{PI}}}\xspace}
\newcommand{\cs}{\textcolor{7621C}{\textbf{\textit{CS}}}\xspace}

\newcommand{\fillpar}{\parfillskip=0pt\par\parfillskip=0pt plus 1fil\relax}

\icmltitlerunning{Envisioning Beyond the Few: Disentangled Semantics and Primitives for Few-Shot Atypical Layout-to-Image Generation}

\begin{document}

\twocolumn[
  \icmltitle{Envisioning Beyond the Few: Disentangled Semantics and Primitives for Few-Shot Atypical Layout-to-Image Generation}



  \icmlsetsymbol{equal}{*}

  \begin{icmlauthorlist}
    \icmlauthor{Nan Bao}{buaa}
    \icmlauthor{Yifan Zhao}{buaa}
    \icmlauthor{Wenzhuang Wang}{buaa}
    \icmlauthor{Jia Li}{buaa}
    \\ \texttt{\{nbao, zhaoyf, wz\_wang, jiali\}@buaa.edu.cn}
  \end{icmlauthorlist}

  \icmlaffiliation{buaa}{State Key Laboratory of Virtual Reality Technology and Systems, School of Computer Science and Engineering and Qingdao Research Institute, Beihang University, China}

  \icmlcorrespondingauthor{Yifan Zhao}{zhaoyf@buaa.edu.cn}
  \icmlcorrespondingauthor{Jia Li}{jiali@buaa.edu.cn}

  \icmlkeywords{layout-to-image generation, few-shot image generation, diffusion models, atypical visual domains}

  \vskip 0.3in
]



\printAffiliationsAndNotice{}  

\begin{abstract}
The layout-to-image (L2I) task enables fine-grained control over image generation via object categories and spatial layouts.
However, existing L2I methods yield fragmented and distorted generations under few-shot atypical settings.
We term this failure as representation fragmentation, arising from a granularity mismatch that entangles semantic identity with visual details.
To address this issue, we propose a representation-driven framework that disentangles semantics from primitives for robust few-shot adaptation.
Specifically, Semantic Anchoring aggregates categorical semantics into anchors for stable identity, while Primitive Imbuing models recomposable primitives for robust local detail modeling.
Conceptual Steering further regulates optimization with a saliency-aware objective to preserve foreground semantic consistency.
Extensive experiments demonstrate consistent improvements in the 5-shot regime over state-of-the-art L2I methods in both visual fidelity and alignment across diverse atypical domains.
The source code is publicly available at \texttt{https://github.com/iCVTEAM/DSP}.
\end{abstract}
\section{Introduction}

\begin{figure*}[!t]
  \vskip 0.2in
  \begin{center}
    \centerline{\includegraphics[width=\linewidth]{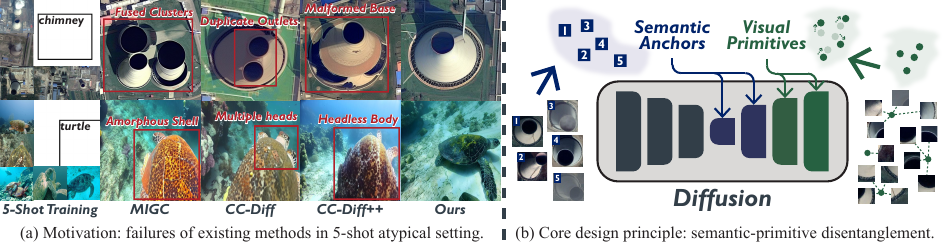}}
    \caption{
\textbf{Few-Shot Atypical L2I via Semantic-Primitive Disentanglement.} 
(a) 
Existing methods suffer from \emph{representation fragmentation}, yielding geometric distortions and fragmented textures (\eg, deformed chimneys and turtles). In contrast, our method maintains structural coherence. 
(b) 
To address the \emph{granularity mismatch} between semantic identity and visual details, we explicitly disentangle representations into \emph{Semantic Anchors} and \emph{Visual Primitives}, further guided by \CS to enforce foreground consistency.\fillpar
}
    \label{fig:Teaser}
  \end{center}
\end{figure*}

The rapid evolution of large-scale diffusion models has revolutionized image generation, unlocking diverse avenues for controllable generation \citep[\eg,][]{compositional, versatilediff, controlnet, controlnet++, freecontrol, t2iadapter}.
Among various controllable paradigms, layout-to-image (L2I) generation stands out as a pivotal bridge between abstract user intent and concrete visual realization, allowing for controllable generation of complex scenes by explicitly defining the category and spatial configurations of constituent objects \citep[\eg,][]{gligen, layoutdiff, instancediff, ficgen, aerogen}.
Despite this success, foundational diffusion models such as Stable Diffusion \cite{ldm} are predominantly trained on large-scale datasets composed of canonical natural images \citep[\eg,][]{laion5b}.
As a result, a pronounced domain gap emerges when applying them to atypical visual domains, such as turbid underwater imagery \cite{ruod} or high-altitude aerial surveillance \cite{dior}, where the underlying visual statistics and geometric priors deviate substantially from the learned training prior.\fillpar

The lack of high-quality data in atypical visual domains further complicates this challenge.
Unlike canonical domains, where paired data is abundant, acquiring and annotating layout-image pairs in atypical conditions is 
labor-intensive. 
Consequently, many real-world applications are inherently constrained to a low-data regime, where only a handful of training samples (\eg, a few shots) are available.
This highlights the critical need for few-shot adaptation methods capable of synthesizing high-fidelity data for specialized domains where collecting large-scale datasets is infeasible.\fillpar

However, directly adapting L2I models to such data-scarce, atypical domains presents \emph{a distinct and underexplored challenge}.
As illustrated in \cref{fig:Teaser} (a), existing L2I methods do not simply memorize the training objects; instead, they often fail to form coherent representations, resulting in fragmented textures and geometric distortions in the generated images.
These failure patterns indicate that, under the few-shot regime, highly variable local visual cues overwhelm global semantics, resulting in fragmented representations.
We term this phenomenon \textbf{representation fragmentation}.\fillpar

To surmount this fragmentation, we argue that robust few-shot atypical L2I generation must fundamentally rethink how representations are formed from limited samples, rather than relying on parameter-level tuning.
Our key insight reveals a critical \textbf{granularity mismatch}: high-level semantic identity demands global consistency and cross-instance stability, while low-level visual textures and appearance details are inherently local and highly variable.
However, existing L2I methods \citep[\eg,][]{ccdiff, ccdiffpp} typically fail to explicitly separate these heterogeneous granularities, resulting in an entangled representation where globally invariant semantics are confounded with stochastic local visual variations.
This confounding underpins the aforementioned fragmentation, where high-level semantic identity, which should remain globally consistent and invariant across instances, is distorted by highly variable local visual details.\fillpar

This observation motivates our core design principle: \emph{explicitly disentangling semantic-level representations from local primitive-level representations} for few-shot atypical L2I, as shown in \cref{fig:Teaser} (b).
To operationalize this principle, we instantiate the disentangled representations through three tightly coupled components, each addressing a specific bottleneck in few-shot adaptation while being jointly optimized as a coherent system.
Specifically, we first stabilize semantic identity across limited exemplars as a prerequisite, then address the ill-posed estimation of fine-grained visual details, and finally enforce semantic fidelity during generation to prevent degenerate shortcuts.
To stabilize semantic identity under the few-shot regime, we introduce \SA, which aggregates categorical semantic evidence across all exemplars into robust semantic anchors.
By emphasizing cross-exemplar consensus rather than exemplar-specific fitting, this design alleviates semantic drift under base-to-novel distribution shift.
Fine-grained visual details, however, is inherently local and poorly constrained in the few-shot regime, necessitating a different approach.
Accordingly, we propose \PI, which models such details as recomposable visual primitives estimated via ridge regression for stable recovery under severe data scarcity.
Finally, to prevent the generative model from circumventing these disentangled representations, we introduce \CS, a saliency-aware objective that explicitly enforces foreground consistency in activation responses.
This component is particularly critical in few-shot transfer due to \textbf{distributional asymmetry}, where background statistics are largely shared while foreground regions undergo substantial shifts.
Together, these components form a representation-driven framework that enforces global semantic consistency for semantic identity while enabling robust recovery of local visual details.
Our contributions are threefold as follows:

1) We highlight representation fragmentation as a dominant failure mode in few-shot generative learning, particularly in atypical L2I settings, and show that mitigating this fragmentation is critical for stable generation under limited inputs.\fillpar

2) We propose a representation-driven framework for few-shot atypical L2I generation, consisting of three tightly coupled components that explicitly disentangle semantic-level representations from local primitive-level representations.\fillpar

3) Extensive experiments demonstrate that our approach consistently outperforms the state-of-the-art L2I methods in fidelity and alignment, achieving robust adaptation with only five training samples across three diverse atypical domains.\fillpar

\section{Related Work}

\subsection{Text-to-Image Generation}

Text-to-image synthesis has evolved from GANs \citep[\eg][]{reed2016generative, xu2018attngan, zhang2021cross, liao2022text} to autoregressive (AR) \cite{dalle, parti, makeascene} and diffusion models \cite{ddpm, imagen}.
While AR methods model images as sequences of discrete tokens, latent diffusion models \cite{ldm} have recently established a new state-of-the-art by performing iterative denoising in a compressed latent space.
Despite this success, relying solely on global text prompts remains insufficient for precise spatial control, which motivates a shift towards layout-to-image paradigms.\fillpar

\subsection{Layout-to-Image Generation}

Layout-to-image (L2I) generation synthesizes images conditioned on structured layouts that specify object instances with category labels and spatial configurations, typically in the form of bounding boxes or segmentation masks.
Recently, a wide range of L2I methods have been proposed.
Most of these methods focus on the challenges of encoding and injecting various types of grounding information, such as different layout representations \cite{blobgen, scpdiff} and instance-level semantic conditions \cite{gligen, reco, layoutdiff, layoutdiffuse, geodiff, migc, migcpp, instancediff, zhang2025creatilayout}, while some address foreground-background correlation issues \cite{ccdiff, ccdiffpp, ficgen}, and a few focus on targeted generation for specific visual scenes or scenarios \cite{aerogen, zhang2025perldiff}.
However, existing approaches predominantly rely on substantial training data, while L2I generation under few-shot regimes remains largely underexplored.\fillpar

\subsection{Few-Shot Learning in Image Generation}

Early few-shot image generation methods primarily relied on GANs \citep[\eg,][]{zhao2022closer}.
With the advent of diffusion models, research has focused on subject-driven personalization, where approaches such as DreamBooth \citep{ruiz2023dreambooth} and DataDream \citep{datadream} fine-tune models using a small set of reference images.
Beyond subject conditioning, adapter-based methods have been proposed to learn novel control signals in low-shot settings \citep{nguyen2025universal}, and domain-specific applications, such as anomaly generation, have explored few-shot diffusion architectures \citep{jin2025dual}.
However, existing few-shot approaches overlook layout-based grounding, leaving few-shot layout-to-image generation an open challenge.\fillpar
\section{Preliminaries}
\label{sec:pre}

\subsection{Latent Diffusion Model} 

Unlike standard DDPMs \cite{ddpm} that operate directly in the high-dimensional pixel domain, Latent Diffusion Models (LDMs) \cite{ldm} perform the generative process within a compressed latent space, thereby achieving significant computational efficiency. 
Consider a dataset $\mathcal{X} = \{(I, y)\}$ containing pairs of images $I$ and labels $y$. 
The input image $I$ is first mapped to a latent code $x_0$ via a pretrained encoder $\mathcal{E}$ of Variational Autoencoder \cite{vae}, formalized as $x_0 \sim q_{\mathcal{E}}(x|I)$.\fillpar

The forward diffusion process is modeled as a Markov chain $\{x_t\}^T_{t=0}$ with Gaussian transitions. 
At any arbitrary timestep $t$, the noisy latent $x_t$ can be sampled directly in closed form via $x_t = \sqrt{\bar\alpha_t}x_0 + \sqrt{1-\bar\alpha_t}\epsilon$, where $\epsilon \sim \mathcal{N}(0, \mathbf{I})$ is the standard Gaussian noise, and $\bar\alpha_t$ defines the noise schedule.\fillpar

The generative process aims to reverse this transition conditioned on the label $y$. 
A denoising U-Net \cite{unet} $\epsilon_\Theta$, parameterized by $\Theta$, is trained to predict the noise $\epsilon$ added to $x_0$. 
The training objective is defined as
\begin{equation}
  \begin{split}
  \min_\Theta \mathcal{L}_{\text{LDM}} = \mathbb{E}_{(I, y) \sim \mathcal{X}, \epsilon, t} \big[\|\epsilon_\Theta(x_t, t, \tau(y)) - \epsilon\|^2_2\big],
  \end{split}
  \label{eq:ldm_obj}
\end{equation}
where $\tau(y)$ maps the label $y$ into a conditional embedding.\fillpar

\subsection{Regime Shift: From Base Priors to Novel}

We formulate the layout-to-image generation task within a few-shot regime.
Let $\mathcal{C}$ denote the set of all semantic categories. 
A layout, serving as the structured label for an image $I_i$, is defined as $y_i = \{ \mathcal{O}_i, g_i \}$. 
It consists of a global caption $g_i$ and a set of $M_i$ objects $\mathcal{O}_i = \{(c_{i,j}, p_{i,j})\}_{j=1}^{M_i}$, 
where $c_{i,j} \in \mathcal{C}$ and $p_{i,j}$ represent the category and bounding box coordinates of the $j$-th object in image $I_i$, respectively.\fillpar

The category set $\mathcal{C}$ is partitioned into two disjoint subsets, base categories $\mathcal{C}_{\text{base}}$ and novel categories $\mathcal{C}_{\text{novel}}$. 
Accordingly, the dataset is decomposed into $\mathcal{X}_{\text{base}} \cup \mathcal{X}_{\text{novel}}$. 
Specifically, $\mathcal{X}_{\text{novel}}$ adheres to a $K$-shot setting, containing exactly $K$ image-layout pairs for each novel category.
We denote this few-shot dataset as $\mathcal{X}_{\text{novel}} = \{(I_i, y_i)\}_{i=1}^{N_{\text{novel}}}$, where $N_{\text{novel}} = |\mathcal{C}_{\text{novel}}| \times K$ is the total number of support images.\fillpar

A two-stage transfer learning paradigm, rather than episodic meta-learning, is adopted to handle this setting.
In the \textbf{base stage}, a conditional latent diffusion model is pretrained on the large-scale dataset $\mathcal{X}_{\text{base}}$ via vanilla cross-attention injection to acquire general layout-to-image generation capabilities, yielding parameters $\Theta_{\text{base}} = \operatorname*{arg\,min}_{\Theta} \mathcal{L}_{\text{LDM}}(\mathcal{X}_{\text{base}}; \Theta)$ optimized using the standard diffusion denoising objective.\fillpar

In the subsequent \textbf{novel stage}, the focus shifts to adapting the model to novel categories using the limited $K$-shot dataset $\mathcal{X}_{\text{novel}}$. 
Initialized with $\Theta_{\text{base}}$, the base model is kept entirely frozen. Only the parameters $\Theta_{\text{novel}}$ of our newly proposed disentanglement modules are updated to approximate the conditional distribution $p_\Theta(I|y)$ for $(I, y) \sim \mathcal{X}_{\text{novel}}$, 
enabling high-fidelity synthesis conditioned on novel layouts.\fillpar

\section{Method}

\begin{figure*}[ht]
  \vskip 0.2in
  \begin{center}
    \centerline{\includegraphics[width=\linewidth]{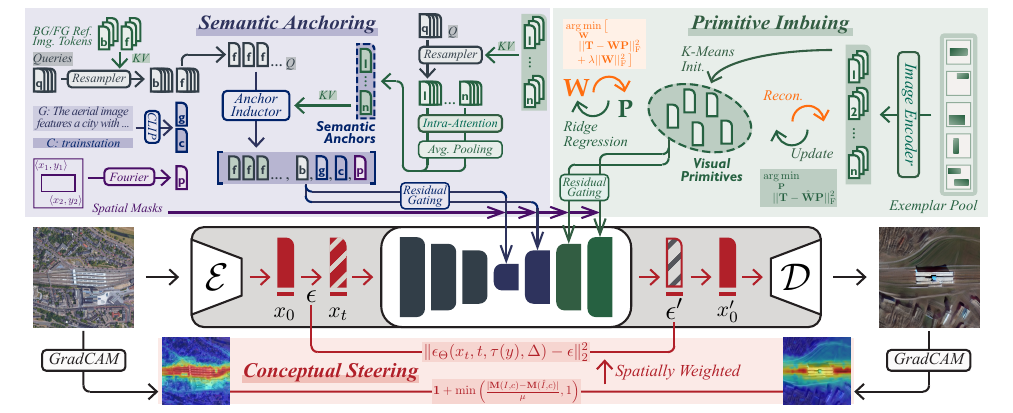}}
    \caption{
\textbf{Method Overview.} 
We propose an atypical few-shot L2I framework comprising:
\SA for categorical semantic stability, \PI for fine-grained local detail recovery, and \CS for saliency-aware foreground optimization.\fillpar
    }
    \label{fig:Main}
  \end{center}
\end{figure*}


Few-shot atypical L2I poses significant challenges, primarily due to representation fragmentation arising from a granularity mismatch that entangles semantic identity with visual details.
To overcome this, we propose a framework incorporating multi-granular control mechanisms, as illustrated in \cref{fig:Main}.
After establishing prior-grounded representations in \cref{sec:init}, we present \SA (\cref{sec:sa}) to stabilize categorical semantic identity.
Complementing this, \PI (\cref{sec:pi}) employs ridge regression optimization to capture fine-grained primitives for robust local modeling.
Furthermore, \CS (\cref{sec:cs}) regulates the optimization by preserving foreground consistency via a saliency-aware objective.\fillpar

\subsection{Prior-Grounded Representation Encoding}
\label{sec:init}

Before delving into the proposed modules, we first formalize the conditional representations used throughout the generation process.
Rather than training from scratch, 
We initialize a set of distinct embeddings by integrating off-the-shelf foundation priors for general semantics and base-learned priors for structural and visual details, ensuring foundational control of layout and semantics over the generation process.\fillpar

\paragraph{Constructing Exemplar Pool.} Given the few-shot dataset $\mathcal{X}_{novel} = \{(I_i, y_i)\}_{i=1}^{N_{\text{novel}}}$, we construct an exemplar pool as
\begin{equation}
\Delta = \bigcup_{i=1}^{N_{\text{novel}}} \{ (\text{crop}(I_i, p_{i,j}), c_{i,j}) \}_{j=1}^{M_i},
\end{equation}
by cropping object instances of each image based on its ground-truth bounding box. This foreground memory bank serves as the source for providing supportive visual details.\fillpar

\paragraph{Initializing Embeddings.} We categorize the input embeddings into three groups based on their information source. For the global caption $g$ and category label $c$, we employ a fixed CLIP \cite{clip} text encoder to obtain the global semantic embedding $\phi_g$ and category embedding $\phi_c$.\fillpar

To incorporate structural layout, we encode the bounding box coordinates using Fourier feature encoding to obtain position embeddings $\phi_p$. 
Additionally, spatial masks $\mathbf{S}$ are derived via a sigmoid function to perform regional selective injection based on the bounding box coordinates of layouts.\fillpar

To capture visual embeddings, we utilize a Resampler module inspired by \cite{perceiver, ccdiff}. 
This module is pre-trained on the $\mathcal{X}_{base}$ and remains frozen during the novel adaptation to ensure feature stability, serving as a generic feature extractor. 
Regarding the input configuration, we consistently retrieve contextually relevant background samples from $\mathcal{X}_{base}$ to extract background embeddings $\phi_b$.
In contrast, for foreground embeddings $\phi_f$, we switch the source from $\mathcal{X}_{base}$ to representative instances sampled from the exemplar pool $\Delta$ for the novel categories.\fillpar

Collectively, these embeddings $\{\phi_f, \phi_b, \phi_g, \phi_c, \phi_p\}$ provide a coarse-grained representation for subsequent adaptation.\fillpar

\subsection{Categorical Semantic Anchoring}
\label{sec:sa}

In the context of few-shot layout-to-image generation, we observe a distinct distributional asymmetry where background statistics remain relatively consistent between the base and novel sets, whereas foreground categories are entirely disjoint.
While freezing the Resampler preserves pre-trained priors, it inherently limits the embeddings to a coarse-grained level lacking precise semantics for the disjoint novel categories.
To bridge this gap, we propose \SA. This mechanism stabilizes categorical perception by establishing semantic anchors derived from the exemplar pool to refine the foreground embeddings $\phi_f$.\fillpar

\paragraph{Retrieving Semantic Anchors.} 
Unlike the general text-derived embeddings $\phi_g$ and $\phi_c$, Semantic Anchors encapsulate crucial image-level semantics derived from scarce novel exemplars.
For a specific category $c$, we retrieve its visual exemplars $\delta_c$ and encode them using a frozen DINOv2 image encoder \cite{dinov2} into dense features $\phi_{(\Delta,c)}\in \mathbb{R}^{n\times h\times w \times d}$, where $n=|\delta_c|$. 
To condense these features, a frozen Resampler queries $\phi_{(\Delta,c)}$ with learnable tokens $\phi_q\in\mathbb{R}^{r,d}$ via cross-attention as
\begin{equation}
\mbox{\small $\displaystyle
\phi_r=\text{softmax}\!\left(\frac{(\phi_q\mathbf{W}_Q^\text{R})(\phi_{(\Delta,c)}\mathbf{W}_K^\text{R})^\top}{\sqrt{d}}\right)(\phi_{(\Delta,c)}\mathbf{W}_V^\text{R}),
$}
\end{equation}
yielding reduced features $\phi_r\in\mathbb{R}^{n\times r\times d}$.
Finally, to capture internal structural dependencies, we obtain the anchors $\mathbf{A} \in \mathbb{R}^{n\times d}$ by averaging the intra-exemplar self-attention outputs along the token dimension $r$, formulated as
\begin{equation}
\mbox{\small $\displaystyle
\mathbf{A} = \frac{1}{r} \sum_{k=1}^{r} \left[ \text{softmax}\!\left(\frac{(\phi_r\mathbf{W}_Q^\text{I})(\phi_r\mathbf{W}_K^\text{I})^\top}{\sqrt{d}}\right) (\phi_r\mathbf{W}_V^\text{I}) \right]_k,
$}
\end{equation}
where $[\cdot]_k$ denotes the $k$-th token vector.
These anchors capture the condensed categorical semantics for each novel exemplar, stabilizing the representation of novel categories.\fillpar

\paragraph{Inducing Semantics with Anchors.}
Since anchors $\mathbf{A}$ are derived exclusively during the novel phase, they require strategic integration into the base pre-training feature stream. Leveraging the similar derivation between $\phi_f$ and $\mathbf{A}$, we integrate the derived anchors with $\phi_f$ to explicitly \textit{induce} categorical semantics, thereby enhancing semantic stability.\fillpar

Formally, we implement this via a gated cross-attention mechanism with a mask bias $\mathcal{M}$ to filter padding tokens. The induced feature $\tilde{\phi}_f$ is formulated as
\begin{equation}
\mbox{\small $\displaystyle
\tilde{\phi}_f = \phi_f + \eta \cdot \text{softmax}\!\left( \frac{(\phi_f\mathbf{W}_Q')(\mathbf{A}\mathbf{W}_K')^\top}{\sqrt{d}} + \mathcal{M} \right) (\mathbf{A}\mathbf{W}_V'),
$}
\end{equation}
where $\mathbf{W}_{\{Q,K,V\}}'$ are projection matrices, and $\eta$ is a learnable gating factor initialized to 0 to preserve pre-trained priors. The bias $\mathcal{M}_{ij}$ is set to $-\infty$ for paddings and $0$ otherwise, ensuring attention is focused solely on valid anchors.\fillpar

The final embedding set $\{\tilde{\phi}_f, \phi_b, \phi_g, \phi_c, \phi_p\}$ is then injected into the \textbf{middle and early upsampling stages} of the U-Net via masked cross-attention. By targeting these structural layers, we ensure the anchors effectively guide the global layout. This injection employs sigmoid-based spatial weighting, following standard conditioning paradigms in latent diffusion models~\cite{gligen, controlnet}.\fillpar

\subsection{Visual Primitives Modeling and Imbuing}
\label{sec:pi}

Although \SA effectively stabilizes categorical semantics, reconstructing fine-grained, highly variable local appearances remains challenging in the few-shot regime. To address this, we propose \PI to complement semantic control with explicit local guidance.\fillpar

\paragraph{Visual Primitives Modeling and Learning.} Inspired by \citet{wertheimer2021few, freelunch}, we model Visual Primitives as a set of compact, learnable prototype embeddings designed to capture visual details. Distinct from the global Semantic Anchors, these primitives operate locally, representing the fundamental geometric fragments.\fillpar

For each target category $c\in \mathcal{C_{\text{novel}}}$, we construct a support set $\delta_c\subset \Delta$ containing $n$ exemplars. 
We first extract dense feature maps $\phi_{(\Delta,c)}\in \mathbb{R}^{n\times h\times w\times d}$ utilizing a frozen DINOv2 image encoder \cite{dinov2}, where $h, w$ represent the spatial resolution and $d$ denotes the channel dimension. 
To establish a holistic representation, these features are subsequently flattened and concatenated along the spatial axes, yielding a unified target matrix $\mathbf{T}\in \mathbb{R}^{nhw\times d}$.\fillpar

Visual Primitives are instantiated as $s$ learnable prototype embeddings $\mathbf{P}\in \mathbb{R}^{s\times d}$, initialized via K-Means clustering on $\mathbf{T}$. To capture local compositionality, we reconstruct $\mathbf{T}$ as a linear combination of these prototypes, \ie, $\mathbf{T} \approx \mathbf{W}\mathbf{P}$.\fillpar

First, fixing the prototypes $\mathbf{P}$, we determine the optimal coefficient matrix $\hat{\mathbf{W}}$ by minimizing the reconstruction error subject to Tikhonov regularization, formulated as
\begin{equation}
    \hat{\mathbf{W}} = \operatorname*{arg\,min}_W \|\mathbf{T}-\mathbf{W}\mathbf{P}\|^2_\text{F} + \lambda\|\mathbf{W}\|^2_\text{F},
\end{equation}
where $\|\cdot\|_\text{F}$ denotes the Frobenius norm and $\lambda$ is a regularization coefficient ensuring numerical stability.
A distinct advantage of this formulation is that it admits a closed-form solution, avoiding the need for computationally expensive iterative solvers. The optimal coefficients are derived as
\begin{equation}
    \hat{\mathbf{W}} = \mathbf{T}\mathbf{P}^\top(\mathbf{P}\mathbf{P}^\top+\lambda\mathbf{I})^{-1}.
\end{equation}

Subsequently, we refine the primitives $\mathbf{P}$ to better align with the target features. By utilizing the derived $\hat{\mathbf{W}}$, we optimize $\mathbf{P}$ by minimizing the reconstruction error, measured by the squared Frobenius norm, formulated as $\|\mathbf{T}-\hat{\mathbf{W}}\mathbf{P}\|^2_\text{F}$.
The complete optimization procedure is detailed in \cref{alg:primitives}.\fillpar

\begin{algorithm}[tb]
   \caption{Alternating Minimization for Primitives}
   \label{alg:primitives}
\begin{algorithmic}[1]
   \STATE {\bfseries Input:} Exemplar subset $\delta_c \subset \Delta$, Number of Primitives $s$, Regularization $\lambda$, Iterations $N_\text{iter}$.
   \STATE {\bfseries Output:} Visual Primitives $\mathbf{P}$, Coefficient Matrix $\mathbf{W}$.
   
   \item[] \hfill $\rhd~$\textcolor{blue}{\textit{Feature Extraction \& Flattening}}
   \STATE Extract features $\phi_{(\Delta,c)} \in \mathbb{R}^{n \times h \times w \times d} \leftarrow \text{Encoder}(\delta_c)$.
   \STATE Flatten $\phi_{(\Delta,c)}$ to obtain target matrix $\mathbf{T} \in \mathbb{R}^{nhw \times d}$.
   
   \item[] \hfill $\rhd~$\textcolor{blue}{\textit{Initialization}}
   \STATE Initialize $\mathbf{P} \in \mathbb{R}^{s \times d}$ via K-Means clustering on $\mathbf{T}$.
   
   \item[] \hfill $\rhd~$\textcolor{blue}{\textit{Optimization Loop}}
   \FOR{$t=1$ {\bfseries to} $N_\text{iter}$}
      \item[] \hfill $\rhd~$\textcolor{blue}{\textit{Step A: Update Coefficients $\mathbf{W}$}}
      \STATE Compute optimal $\mathbf{W}$ with $\mathbf{P}$ fixed:
      \item[] \quad $\mathbf{W} \leftarrow \mathbf{T}\mathbf{P}^\top(\mathbf{P}\mathbf{P}^\top + \lambda\mathbf{I})^{-1}$
      
      \item[] \hfill $\rhd~$\textcolor{blue}{\textit{Step B: Update Primitives $\mathbf{P}$}}
      \STATE Update $\mathbf{P}$ by minimizing recon. error with $\mathbf{W}$ fixed:
      \item[] \quad $\mathbf{P} \leftarrow \operatorname*{arg\,min}_{\mathbf{P}} \|\mathbf{T} - \mathbf{W}\mathbf{P}\|_\text{F}^2$
   \ENDFOR
   
   \STATE \textbf{return} $\mathbf{P}, \mathbf{W}$
\end{algorithmic}
\end{algorithm}

\paragraph{Imbuing Primitives into Generation.}
Given the learned visual primitives $\mathbf{P}$, we aim to imbue them into the generation process.
To specifically enhance the synthesis of fine-grained visual details, we target the \textbf{later and final upsampling stages} of the U-Net.
Let $\mathbf{h}$ denote the intermediate spatial features within a cross-attention layer. 
We inject $\mathbf{P}$ into $\mathbf{h}$ via a spatial-gated mechanism defined as
\begin{equation}
\mbox{\small $\displaystyle
\tilde{\mathbf{h}} = \mathbf{h} + \gamma \cdot \mathcal{G} \odot
\operatorname{softmax}\!\left(
\frac{(\mathbf{h}\mathbf{W}_Q'')(\mathbf{P}\mathbf{W}_K'')^\top}{\sqrt{d}}
\right) (\mathbf{P}\mathbf{W}_V''),
$}
\end{equation}
where $\mathbf{W}_{Q,K,V}''$ are projection matrices, and $\gamma$ is a learnable scalar initialized to zero.
Here, $\mathcal{G}$ serves as a sparsity-enforced spatial gate derived from the layout bounding boxes.
Specifically, we define $\mathcal{G} = \mathbf{S} \odot \mathbf{1}_{\text{top}}(\mathbf{S})$, where $\mathbf{S}$ denotes the dense sigmoid-based spatial masks and $\mathbf{1}_{\text{top}}(\cdot)$ acts as a hard selection operator preserving only the most significant activations. 
This formulation strictly confines the primitive injection to the most salient foreground regions.\fillpar

\subsection{Saliency-Aware Conceptual Steering}
\label{sec:cs}

In few-shot adaptation, the disentangled representations are not guaranteed to be effectively utilized during generation.
Due to the pronounced distributional asymmetry between foreground and background, background regions typically exhibit lower variability and more stable gradients under standard diffusion training.
As a result, optimization tends to reduce reconstruction error by preferentially fitting background patterns that are weakly coupled with the novel concepts, thereby weakening the semantic fidelity in the target objects.
We therefore introduce \CS, which explicitly enforces foreground-focused optimization.\fillpar

To implement this, we employ text-driven GradCAM \cite{gradcam, clipgradcam} as a semantic locator. By measuring the divergence in activation maps between the generated outputs and the ground truth, this module identifies specific regions of semantic inconsistency, thereby guiding the model toward more precise concept alignment.\fillpar

Formally, focusing on the target category $c$ specified in the layout $y$, we define a spatial penalty mask $\boldsymbol{\Omega}$. 
This mask quantifies the discrepancy between the activation maps $\mathbf{M}(\cdot, c)$ of the ground-truth image $I$ and the predicted original image $\hat{I}$ reconstructed from $x_t$, formulated as
\begin{equation}
  \boldsymbol{\Omega} = \mathbf{1} + \min\left( \frac{|\mathbf{M}(I, c) - \mathbf{M}(\hat{I}, c)|}{\mu}, 1 \right),
\end{equation}
where $\mu$ is a sensitivity threshold.
We incorporate this mask into the objective to explicitly penalize misalignment as
\begin{equation}
  \mathcal{L}_{\text{final}} = \mathbb{E}_{(I, y) \sim \mathcal{X}_\text{novel}, \epsilon, t} \Big[ \big\| \boldsymbol{\Omega} \odot \big(\epsilon_\Theta(x_t, t, \tau(y), \Delta) - \epsilon\big) \big\|^2_2 \Big],
  \label{eq:cs_loss}
\end{equation}
where $\odot$ denotes the element-wise Hadamard product.
By spatially modulating the loss, $\boldsymbol{\Omega}$ effectively amplifies gradients in under-activated regions, thereby ensuring the novel concepts are faithfully imprinted onto the intended layouts.\fillpar
\section{Experiments}

\begin{figure*}[!t]
  \vskip 0.2in
  \begin{center}
    \centerline{\includegraphics[width=\linewidth]{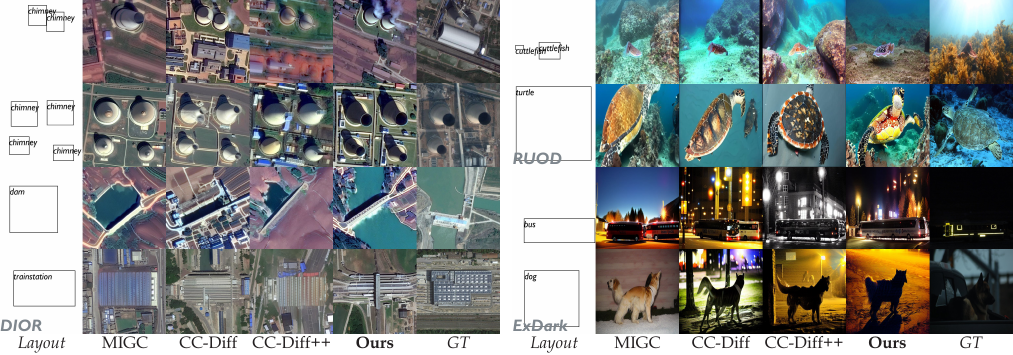}}
    \caption{
Qualitative comparisons under the 5-shot setting on aerial, underwater, and extreme dark domains.
    }
    \label{fig:Exp}
  \end{center}
\end{figure*}

\subsection{Experimental Setups}

\paragraph{Datasets.}
We conduct few-shot experiments on three atypical datasets: DIOR \cite{dior} for the aerial domain, RUOD \cite{ruod} for the underwater domain, and ExDark \cite{exdark} for the low-light domain.
Each dataset is adapted for a few-shot setting by dividing it into base and novel sets based on the foreground categories, while maintaining the original training and testing partitions.
Detailed dataset configurations are provided in \cref{sec:appendix_datasets}.\fillpar

\paragraph{Implementation Details.}
Built upon the Stable Diffusion v1.5 checkpoint, our framework applies \SA to the middle and first upsampling blocks with cross-attention, and \PI to the final two.
The model is trained on four NVIDIA RTX 4090 GPUs using gradient accumulation.
All layout-image pairs for both training and inference are resized to 512 \texttimes~512.
We utilize AdamW \cite{adamw} with a base learning rate of 1 \texttimes~10\textsuperscript{-4}, applying a 100 \texttimes~multiplier to gating parameters $\eta$ and $\gamma$.
\textbf{During the base training phase,} we train the model on the base set for 100 epochs with a total batch size of 320.
\textbf{For few-shot adaptation,} we fine-tune the model for 100 steps. In each step, we perform a full-batch update by accumulating gradients from the entire set of novel samples.
\textbf{For inference,} we employ the Euler Discrete Scheduler \cite{euler} with 50 denoising steps and the classifier-free guidance \cite{clsfree} with a scale of 7.5.
We set $\lambda$ and $\mu$ to 0.1 and 0.95, respectively.
\textbf{Regarding primitive optimization,} alternating minimization is conducted as a standalone process preceding the novel phase fine-tuning. We set the number of iterations $N_\text{iter}$ to 50 and the number of primitives $s$ to 128 to achieve a balanced performance.\fillpar

\paragraph{Evaluation Protocol.}
We compare our method with current SOTAs \cite{migc, ccdiff, ccdiffpp} under a 5-shot setting.
To ensure a rigorous and fair comparison, we utilize a fixed sequence of 50 random seeds throughout the training and evaluation.
Each seed deterministically controls the data sampling process by selecting the 5 support images for fine-tuning and determining the input layouts for generation.
This guarantees that all methods are evaluated on identical data splits and inference conditions.\fillpar

\paragraph{Metrics.} 
We evaluate fidelity using FID \cite{fid} and alignment using the 
COCO evaluation protocol \cite{coco}.
\textbf{Bootstrap FID. }
Standard FID is heavily biased given the small sample sizes in our setting (\ie, 50 images per category), as they are insufficient to accurately model the feature manifold. To address this, we utilize a bootstrapped FID strategy. We pool generations from all 50 seeds and report the mean FID calculated via bootstrap sampling against the entire novel test set over 50 iterations.\fillpar

\subsection{Comparisons with state-of-the-art}

\subsubsection{Quantitative Comparisons}

\begin{table}[t]
  \caption{Quantitative comparison under the 5-shot setting on multiple atypical datasets. mAP, AP\textsubscript{50}, and AP\textsubscript{75} are evaluated by pretrained Faster R-CNN \cite{fasterrcnn} detectors.}
  \label{tab:comp1}
  \begin{center}
    \begin{small}
      \begin{sc}
        \setlength\tabcolsep{2pt}
        \renewcommand{\arraystretch}{1}
        \resizebox{1\columnwidth}{!}{
\begin{tabular}{lcccc}
  \toprule
  Method & FID\textsubscript{Boot} $\downarrow$ & mAP (\%) $\uparrow$ & AP\textsubscript{50} (\%) $\uparrow$ & AP\textsubscript{75} (\%) $\uparrow$ \\
  \midrule
  \multicolumn{5}{c}{\cellcolor{CoolGray1C} DIOR \cite{dior}} \\
   MIGC~\citeyearpar{migc} & 89.20$\pm$ 0.93 & 22.75$\pm$ 0.77 & 51.73$\pm$ 1.35 & 16.14$\pm$ 0.98 \\
   CC-Diff~\citeyearpar{ccdiff} & \second{82.51$\pm$ 0.86} & \second{24.91$\pm$ 0.77} & \second{55.27$\pm$ 1.40} & \second{19.21$\pm$ 0.97} \\
   CC-Diff++~\citeyearpar{ccdiffpp} & 82.62$\pm$ 0.95 & 24.63$\pm$ 0.84 & 54.60$\pm$ 1.44 & 18.71$\pm$ 1.08 \\
   \textbf{Ours} & \first{74.34$\pm$ 0.95} & \first{26.06$\pm$ 0.89} & \first{57.22$\pm$ 1.30} & \first{20.46$\pm$ 1.15} \\
  \midrule
  \multicolumn{5}{c}{\cellcolor{CoolGray1C} RUOD \cite{ruod}} \\ 
   MIGC~\citeyearpar{migc} & 49.25$\pm$ 0.48 & 17.12$\pm$ 0.74 & 38.56$\pm$ 1.53 & 13.27$\pm$ 0.80\\
   CC-Diff~\citeyearpar{ccdiff} & 48.11$\pm$ 0.53 & \second{18.49$\pm$ 0.74} & 40.86$\pm$ 1.47 & \second{14.65$\pm$ 0.92} \\
   CC-Diff++~\citeyearpar{ccdiffpp} & \second{46.46$\pm$ 0.56} & 18.37$\pm$ 0.79 & \second{42.11$\pm$ 1.58} & 13.63$\pm$ 0.81 \\
   \textbf{Ours} & \first{45.44$\pm$ 0.47} & \first{19.45$\pm$ 0.75} & \first{43.43$\pm$ 1.36} & \first{15.12$\pm$ 0.92} \\
  \midrule
  \multicolumn{5}{c}{\cellcolor{CoolGray1C} ExDark \cite{exdark}} \\ 
   MIGC~\citeyearpar{migc} & 93.60$\pm$ 0.86 & 32.88$\pm$ 0.90 & 67.78$\pm$ 1.11 & 28.07$\pm$ 1.51 \\
   CC-Diff~\citeyearpar{ccdiff} & 94.15$\pm$ 0.79 & 35.26$\pm$ 0.82 & 70.19$\pm$ 1.12 & \second{32.08$\pm$ 1.30} \\
   CC-Diff++~\citeyearpar{ccdiffpp} & \second{93.09$\pm$ 0.72} & \second{35.34$\pm$ 1.02} & \second{70.57$\pm$ 1.31} & 31.71$\pm$ 1.44 \\
   \textbf{Ours} & \first{91.36$\pm$ 0.64} & \first{35.93$\pm$ 0.89} & \first{70.99$\pm$ 1.09} & \first{33.13$\pm$ 1.36} \\
  \bottomrule
\end{tabular}
        }
      \end{sc}
    \end{small}
  \end{center}
  \vskip -0.1in
\end{table}
\begin{table}[t]
  \caption{Quantitative comparison under the 5-shot setting on DIOR, evaluated by a pretrained YOLOv8 \cite{yolov8} detector.}
  \label{tab:comp2}
  \begin{center}
    \begin{small}
      \begin{sc}
        \setlength\tabcolsep{2pt}
        \renewcommand{\arraystretch}{1}
        \resizebox{1\columnwidth}{!}{
\begin{tabular}{lcccc}
  \toprule
  \multirow{2}{*}{Method}& \multirow{2}{*}{FID\textsubscript{Boot} $\downarrow$} & \multicolumn{3}{c}{YOLOv8 Detector} \\
   & &  mAP (\%) $\uparrow$ & AP\textsubscript{50} (\%) $\uparrow$ & AP\textsubscript{75} (\%) $\uparrow$ \\
  \midrule
   MIGC~\citeyearpar{migc} & 89.20$\pm$ 0.93 & 16.21$\pm$ 0.87 & 35.72$\pm$ 1.77 & 12.03$\pm$ 1.01 \\
   CC-Diff~\citeyearpar{ccdiff} & \second{82.51$\pm$ 0.86} & 19.50$\pm$ 1.02 & 40.32$\pm$ 1.86 & \second{16.80$\pm$ 1.14} \\
   CC-Diff++~\citeyearpar{ccdiffpp} & 82.62$\pm$ 0.95 & \second{19.50$\pm$ 1.01} & \second{41.20$\pm$ 1.97} & 16.49$\pm$ 1.10 \\
   \textbf{Ours} & \first{74.34$\pm$ 0.95} & \first{20.80$\pm$ 0.99} & \first{43.34$\pm$ 1.68} & \first{17.66$\pm$ 1.21} \\
  \bottomrule
\end{tabular}
        }
      \end{sc}
    \end{small}
  \end{center}
  \vskip -0.1in
\end{table}
\begin{table}[t]
  \caption{Ablation study on each component on DIOR.}
  \label{tab:abla1}
  \begin{center}
    \begin{small}
      \begin{sc}
        \setlength\tabcolsep{4pt}
        \renewcommand{\arraystretch}{1}
        \resizebox{1\columnwidth}{!}{
\begin{tabular}{ccc|ccccc}
  \toprule
   \sa & \pI & \cs & FID\textsubscript{Boot} $\downarrow$  & mAP (\%) $\uparrow$ & AP\textsubscript{50} (\%) $\uparrow$ & AP\textsubscript{75} (\%) $\uparrow$ \\
  \midrule
   & & & 94.96$\pm$ 0.88 & 19.15$\pm$ 0.78 & 47.97$\pm$ 1.33 & 11.13$\pm$ 0.99 \\
   $\surd$ & & & 88.57$\pm$ 0.95 & 22.84$\pm$ 0.96 & 52.05$\pm$ 1.64 & 16.80$\pm$ 1.18 \\
   & $\surd$ & & 87.34$\pm$ 0.80 & 21.09$\pm$ 0.65 & 51.12$\pm$ 1.29 & 12.94$\pm$ 0.76 \\
   $\surd$ & $\surd$ & & 85.00$\pm$ 0.88 & 25.28$\pm$ 0.91 & 56.15$\pm$ 1.56 & 19.57$\pm$ 1.06 \\
   $\surd$ & $\surd$ & $\surd$ & \first{74.34$\pm$ 0.95} & \first{26.06$\pm$ 0.89} & \first{57.22$\pm$ 1.30} & \first{20.46$\pm$ 1.15} \\
  \bottomrule
\end{tabular}
        }
      \end{sc}
    \end{small}
  \end{center}
  \vskip -0.1in
\end{table}
\begin{table}[t]
  \caption{Ablation study on different variants on DIOR.}
  \label{tab:abla2}
  \begin{center}
    \begin{small}
      \begin{sc}
        \setlength\tabcolsep{2pt}
        \renewcommand{\arraystretch}{1}
        \resizebox{1\columnwidth}{!}{
\begin{tabular}{lcccc}
  \toprule
   Variants & FID\textsubscript{Boot} $\downarrow$  & mAP (\%) $\uparrow$ & AP\textsubscript{50} (\%) $\uparrow$ & AP\textsubscript{75} (\%) $\uparrow$ \\
  \midrule
  Reduced \sa Injection & 81.46$\pm$ 0.76 & 17.31$\pm$ 0.59 & 44.92$\pm$ 1.21 & 9.17$\pm$ 0.65 \\
  Reduced \pI Injection & 76.48$\pm$ 0.80 & 25.70$\pm$ 0.83 & 56.60$\pm$ 1.35 & 19.84$\pm$ 1.18 \\
  \pI-\sa Swapping & 83.09$\pm$ 0.65 & 11.89$\pm$ 0.81 & 30.77$\pm$ 1.77 & 6.41$\pm$ 0.72 \\
  Primitives Random Init. & 75.08$\pm$ 0.92 & 25.87$\pm$ 0.86 & 57.83$\pm$ 1.35 & 19.06$\pm$ 1.17 \\
  Primitives Fixed K-Means & 78.23$\pm$ 0.97 & 25.48$\pm$ 0.73 & 56.33$\pm$ 1.16 & 19.35$\pm$ 1.04 \\
  \textbf{Ours (Full Model)} & \first{74.34$\pm$ 0.95} & \first{26.06$\pm$ 0.89} & \first{57.22$\pm$ 1.30} & \first{20.46$\pm$ 1.15} \\
  \bottomrule
\end{tabular}
        }
      \end{sc}
    \end{small}
  \end{center}
  \vskip -0.1in
\end{table}
\begin{table}[t]
  \caption{Ablation study on the number of visual primitives.}
  \label{tab:abla3}
  \begin{center}
    \begin{small}
      \begin{sc}
        \setlength\tabcolsep{8pt}
        \renewcommand{\arraystretch}{1}
        \resizebox{1\columnwidth}{!}{
\begin{tabular}{lcccc}
  \toprule
   $s$ & FID\textsubscript{Boot} $\downarrow$  & mAP (\%) $\uparrow$ & AP\textsubscript{50} (\%) $\uparrow$ & AP\textsubscript{75} (\%) $\uparrow$ \\
  \midrule
   32 & 76.38$\pm$ 0.91 & 25.95$\pm$ 0.79 & \second{57.34$\pm$ 1.29} & 19.72$\pm$ 1.17 \\
   64 & \second{76.03$\pm$ 1.00}  & 25.51$\pm$ 0.85 & 56.84$\pm$ 1.21 & 19.13$\pm$ 1.26 \\
   128 & \first{74.34$\pm$ 0.95} & \second{26.06$\pm$ 0.89} & 57.22$\pm$ 1.30 & \first{20.46$\pm$ 1.15} \\
   256 & 76.30$\pm$ 0.98 & \first{26.26$\pm$ 0.75} & \first{57.41$\pm$ 1.14} & \second{20.12$\pm$ 1.05} \\
  \bottomrule
\end{tabular}
        }
      \end{sc}
    \end{small}
  \end{center}
  \vskip -0.1in
\end{table}
\begin{figure}[t]
  \vskip 0.2in
  \begin{center}
    \centerline{\includegraphics[width=\columnwidth]{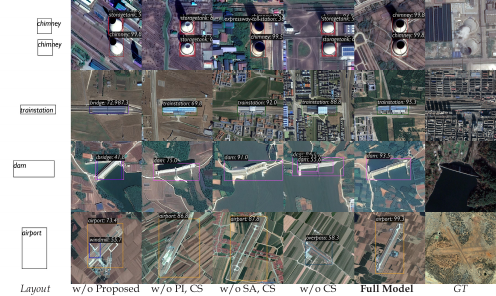}}
    \caption{
    Visualization results of core components ablation.
    }
    \vspace{-0.25em}
    \label{fig:Viz1}
  \end{center}
\end{figure}
\begin{figure}[t]
  \vskip 0.2in
  \begin{center}
    \centerline{\includegraphics[width=\columnwidth]{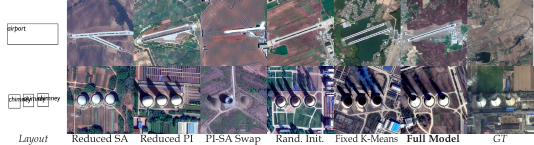}}
    \caption{
    Visualization results of multiple variants.
    }
    \label{fig:Viz2}
  \end{center}
\end{figure}

\Cref{tab:comp1} tabulates the comparative results under the 5-shot setting in terms of fidelity and alignment across aerial \citep[DIOR,][]{dior}, underwater \citep[RUOD,][]{ruod}, and low-light \citep[ExDark,][]{exdark} domains.
Compared with MIGC \cite{migc}, CC-Diff \cite{ccdiff}, and CC-Diff++ \cite{ccdiffpp}, our method consistently achieves strong performance across all evaluated benchmarks.
Notably, our approach yields a substantial improvement on the aerial domain, \textbf{reducing the Bootstrap FID by a significant margin of 8.17} compared to CC-Diff (74.34 vs. 82.51). 
This pronounced gain indicates the effectiveness of our disentangled design in enhancing high-fidelity detail generation under limited data availability.
In addition, regarding layout alignment, our method outperforms all competing approaches across all detection metrics (mAP, AP\textsubscript{50}, and AP\textsubscript{75}).
Such consistent improvements suggest that the generated images exhibit stronger semantic coherence and are more readily recognized by pretrained Faster R-CNN \cite{fasterrcnn} detectors.
Overall, the results across multiple datasets demonstrate the robustness and generalization capability of our proposed method across a diverse range of atypical domains under the 5-shot setting.\fillpar

To further examine the generalization across different detectors, we conduct additional evaluation using a pretrained YOLOv8 \cite{yolov8} on the aerial domain. 
As shown in \Cref{tab:comp2}, our method consistently outperforms CC-Diff and CC-Diff++ across all detection metrics under the 5-shot setting, achieving improvements of 1.30\% in mAP (over both CC-Diff and CC-Diff++), 2.14\% in AP\textsubscript{50} (over CC-Diff++), and 0.86\% in AP\textsubscript{75} (over CC-Diff). 
These results demonstrate that our disentanglement strategy yields detector-agnostic improvements, enabling a more consistent semantical alignment across various detection backbones.\fillpar

\subsubsection{Qualitative Comparisons}

As shown in \cref{fig:Exp}, our method demonstrates superior performance across diverse scenarios. In the aerial domain, it generates structurally stable chimneys, a contextually coherent dam surrounded by water, and a train station with richer texture details compared to competing methods. In the underwater domain, our method successfully generates a small cuttlefish where others fail. It also generates a realistic turtle, avoiding the anatomical errors found in competing methods, such as misplaced fins on the head of CC-Diff or missing fins of CC-Diff++. Finally, in the extreme dark domain, our method maintains object integrity by generating a single, coherent bus, whereas MIGC and CC-Diff fragment the object. We also accurately reconstruct the dog's tail and leg count, eliminating the structural anomalies (\eg, six-legged dogs in CC-Diff) observed in competing methods.\fillpar

\subsection{Ablation Studies and Analyses}

\begin{figure}[t]
  \vskip 0.2in
  \begin{center}
    \centerline{\includegraphics[width=\columnwidth]{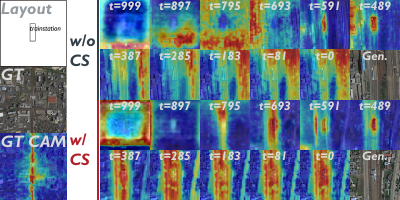}}
    \caption{
    Visualization results of Class Activation Maps (CAMs) with and without \cs across varying inference timesteps $t$.
    }
    \vspace{-0.25em}
    \label{fig:Viz}
  \end{center}
\end{figure}
\begin{figure}[t]
  \vskip 0.2in
  \begin{center}
    \centerline{\includegraphics[width=\columnwidth]{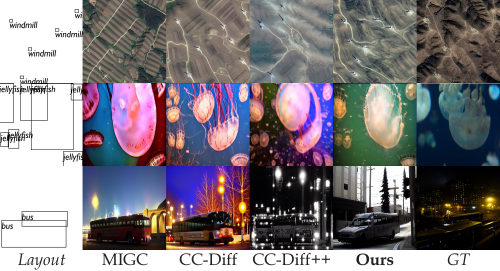}}
    \caption{
    Visualization results of failure cases.
    }
    \label{fig:VizFc}
  \end{center}
\end{figure}

\paragraph{Ablation on Core Components.} 
We first examine the contribution of each core component, as shown in \Cref{tab:abla1}.
Enabling either \sa or \pI individually improves both fidelity and layout alignment over the baseline, while their combination leads to a substantial performance gain,  demonstrating a strong synergistic effect between the two modules.
Introducing \cs on top of \sa and \pI yields a significant improvement in fidelity, indicating its effectiveness in stabilizing training and reinforcing global consistency.
\Cref{fig:Viz1} further confirms that the absence of these components compromises both bounding box localization and detection confidence.\fillpar

\paragraph{Ablation on Multiple Variants.} 
We further analyze several design choices in \Cref{tab:abla2}.
Reducing the number of injection layers for either \sa or \pI leads to clear performance degradation.
When \sa is weakened while \pI and \cs remain active, the resulting performance drop becomes more pronounced, indicating that these modules are not independently effective but rely on a well-balanced synergistic configuration.
Swapping the roles of \sa and \pI leads to a severe performance collapse across all metrics, suggesting that each module fulfills a distinct role within the designed representation hierarchy and that violating this assignment breaks the module synergy.
Finally, we evaluate different primitive initialization strategies. Both random initialization and fixed K-means primitives underperform the full model, highlighting the benefit of our ridge-regression-based optimization strategy in learning more effective primitives.
\Cref{fig:Viz2} confirms the quality drop, especially where the \pI-\sa swap confuses the airport with an airplane structure.\fillpar

\paragraph{Ablation on Primitive Numbers.}
\Cref{tab:abla3} reports the performance under varying primitive numbers.
We observe that $s$ = 128 yields the optimal balance, whereas overly small or large primitive counts slightly degrade fidelity and alignment, suggesting a balanced primitive capacity is crucial.\fillpar

\paragraph{Qualitative Analysis of CS.}
To analyze the impact of \cs, we visualize CAMs at different inference steps $t$, as shown in \cref{fig:Viz}. 
Without \cs, CAMs rarely cover the layout elements. 
However, with \cs, CAMs accurately highlight these regions from $t \approx 800$. 
This validates that our method ensures a better semantical consistency during generation.\fillpar

\paragraph{Failure Case Analysis.}
As shown in \cref{fig:VizFc}, the primary failure mode is the object count mismatch, which predominantly occurs with extremely small objects (\eg, windmills) or complex overlapping boxes (\eg, jellyfish and buses). Under these challenging conditions, existing competing methods exhibit the same limitations, highlighting the need for more robust counting and occlusion modeling.\fillpar
\section{Discussion and Limitations}

\label{sec:limitations}

First, our framework inherits the spatial resolution bottleneck of UNet-based models. Operating within a compressed latent space makes conditioning tiny spatial regions fundamentally challenging, as features for minute objects can collapse into less than a single pixel. While our disentangled design enhances fine-grained control, it cannot entirely bypass this physical encoder limitation, leading to count mismatches and geometric distortions under extreme scales.\fillpar

Additionally, our framework utilizes pre-trained external modules like DINOv2 \cite{dinov2} and Grad-CAM \cite{clipgradcam}. While leveraging such external priors is a pragmatic practice to mitigate domain shift under extreme data scarcity, it introduces prior knowledge that limits the overall self-sufficiency of the approach. Future research could explore more self-contained disentanglement mechanisms that exclusively rely on the internal representations of the diffusion model to further reduce these dependencies.\fillpar

Finally, our empirical validation focuses on the UNet-based Stable Diffusion v1.5 rather than recent Diffusion Transformers (DiTs) \cite{DiT, SD3, flux2024}. Our primary objective is to rigorously verify the proposed representation disentanglement mechanism itself. SD1.5 serves as a transparent testbed where its mature cross-attention mechanism clearly isolates the contribution of our approach. Furthermore, its manageable scale enables the extensive multi-run fine-tuning necessary for statistically reliable few-shot validation. In contrast, fine-tuning 10B+ parameter DiTs remains computationally prohibitive. 
While our framework operates at the representation level and shows potential compatibility with Transformer backbones, extending it to DiTs requires redesigning spatial conditioning mechanisms to align with distinct sequence-based attention structures. We leave this architectural adaptation and the associated retraining requirements for future work.\fillpar

\section{Conclusion}

In this paper, we identify representation fragmentation as the primary bottleneck in few-shot atypical L2I generation. 
To address this, we propose a representation-driven framework that resolves granularity mismatches by disentangling semantics from visual primitives. 
Extensive experiments demonstrate that our method consistently outperforms existing L2I approaches in fidelity and alignment under the 5-shot setting across various atypical domains. 
These results underscore the effectiveness of our disentanglement strategy in stabilizing generation for data-scarce atypical scenarios.\fillpar

\section*{Acknowledgement}

This work is partially supported by grants from the National Natural Science Foundation of China (No. 62132002), the Guizhou Provincial Major Scientific and Technological Program (Qiankehe Zhongda [2025] No. 032), the Beijing Nova Program (No. 20250484786), and the Fundamental Research Funds for the Central Universities.

\section*{Impact Statement}

This paper presents work aimed at improving few-shot image generation in specialized domains such as aerial and underwater imagery. 
Potential positive impacts include alleviating data scarcity and reducing data collection costs in these fields.
As our method is built upon existing generative frameworks, it may inherit the standard risks associated with generative models, including the production of unrealistic or misleading visual content.
However, since this work focuses on restricted domain-specific scenarios rather than open-domain image synthesis, we believe it does not raise fundamentally novel ethical concerns beyond those commonly associated with generative modeling research.

\bibliography{main}
\bibliographystyle{icml2026}

\newpage
\appendix
\onecolumn
\section*{Appendix}

\newcommand{\cls}[1]{\textit{#1}}

\section{Dataset Configurations}
\label{sec:appendix_datasets}

In this section, we provide a comprehensive description of how the DIOR, RUOD, and ExDark datasets were adapted for the few-shot layout-to-image generation task. While we strictly adhered to the original training and testing splits of each benchmark to ensure fair comparison, we reorganized the internal data distribution based on the semantic categories of the foreground objects. The specific partition strategies and data curation protocols are detailed below.

\subsection{Semantic Category Partitioning}
To facilitate few-shot learning, we partitioned the label space of each dataset into disjoint \textit{base} and \textit{novel} sets ($\mathcal{C}_\text{base}$ and $\mathcal{C}_\text{novel}$). 

For the \textbf{DIOR} dataset, the semantic space is divided such that $\mathcal{C}_\text{base}$ comprises 15 categories, including \cls{vehicle}, \cls{baseballfield}, \cls{groundtrackfield}, \cls{bridge}, \cls{overpass}, \cls{ship}, \cls{airplane}, \cls{tenniscourt}, \cls{expressway-service-area}, \cls{basketballcourt}, \cls{stadium}, \cls{storagetank}, \cls{expressway-toll-station}, \cls{golffield}, and \cls{harbor}. Conversely, the $\mathcal{C}_\text{novel}$ set consists of \cls{windmill}, \cls{airport}, \cls{chimney}, \cls{dam}, and \cls{trainstation}. 

Regarding the \textbf{RUOD} dataset, we designate \cls{holothurian}, \cls{echinus}, \cls{scallop}, \cls{starfish}, \cls{fish}, and \cls{diver} as the base categories, while reserving \cls{corals}, \cls{cuttlefish}, \cls{turtle}, and \cls{jellyfish} as the novel categories for few-shot adaptation. 

Finally, for the \textbf{ExDark} dataset, the category split assigns \cls{bicycle}, \cls{boat}, \cls{bottle}, \cls{car}, \cls{cat}, \cls{chair}, \cls{cup}, and \cls{people} to the base set, with \cls{bus}, \cls{dog}, \cls{motorbike}, and \cls{table} constituting the novel set.

\begin{table}[h]
\caption{Statistics of the adapted datasets after filtering. We report the number of images available for the \textit{base} (aggregated) and \textit{novel} (per-category) splits in both the training and testing sets. Note that images containing multiple distinct novel categories were excluded.}
\label{tab:dataset_stats}
\begin{center}
\begin{small}
\begin{sc}

\setlength\tabcolsep{28pt}
\renewcommand{\arraystretch}{1}
\resizebox{1\linewidth}{!}{
\begin{tabular}{lllrr}
\toprule
Dataset & Split Type & Category & \# Train & \# Test \\
\midrule
\multirow{6}{*}{DIOR} 
 & Base (15 classes) & Aggregated & 4449 & N/A \\
 \cmidrule(l){2-5} 
 & \multirow{5}{*}{Novel} 
   & Airport       & 241 & 447 \\
 & & Chimney       & 121 & 256 \\
 & & Dam           & 203 & 379 \\
 & & Trainstation  & 160 & 332 \\
 & & Windmill      & 403 & 803 \\
\midrule
\multirow{5}{*}{RUOD} 
 & Base (6 classes) & Aggregated & 4419 & N/A \\
 \cmidrule(l){2-5} 
 & \multirow{4}{*}{Novel} 
   & Corals        & 138 & 58 \\
 & & Cuttlefish    & 949 & 394 \\
 & & Jellyfish     & 342 & 124 \\
 & & Turtle        & 743 & 309 \\
\midrule
\multirow{5}{*}{ExDark} 
 & Base (8 classes) & Aggregated & 1740 & N/A \\
 \cmidrule(l){2-5} 
 & \multirow{4}{*}{Novel} 
   & Bus           & 95 & 99\\
 & & Dog           & 135 & 383 \\
 & & Motorbike     & 62 & 68 \\
 & & Table         & 58 & 54 \\
\bottomrule
\end{tabular}
}
\end{sc}
\end{small}
\end{center}
\end{table}

\subsection{Data Curation and Protocol}
To ensure a rigorous evaluation setting where the definition of ``$K$-shot'' is statistically precise, we implemented a strict filtering strategy dependent on the co-occurrence of object categories within a single image. 

First, images containing exclusively base category objects were retained to construct the base training set, which is utilized during the initial \textbf{base training} phase. 
Second, to prevent information leakage and ensure a clear separation between known and unknown concepts, any images containing a mixture of both base and novel foreground objects were \textbf{systematically discarded}. 
Third, we excluded images containing objects from \textbf{multiple different} novel categories simultaneously. This exclusion is necessary because the presence of overlapping novel categories renders the $K$-shot statistics ill-defined; specifically, it becomes impossible to uniquely assign such an image to the discrete instance count of a single category, thereby complicating the precise tracking of the $K$-shot budget. 
Note that images containing multiple instances of the \textit{same} novel category are retained.

During the \textbf{novel fine-tuning} phase, we sample the training data from the filtered novel candidates using specific random seeds, strictly selecting exactly $K$ images per category to adhere to the $K$-shot constraint. 
Subsequently, during the \textbf{inference} phase, to maintain computational feasibility without sacrificing statistical significance, we randomly sample 50 layout conditions for each novel category from the novel test set, again controlled by specific seeds to ensure reproducibility. 
The quantitative statistics resulting from this reorganization are summarized in \cref{tab:dataset_stats}.

\section{Detailed Experimental Configurations}

To ensure full transparency and a fair comparative environment, we systematically document the fine-tuning mechanics and hyperparameter configurations applied to our proposed method and the evaluated baselines (MIGC, CC-Diff, and CC-Diff++). We deliberately avoided baseline-specific hyperparameter tuning to maintain a rigorously controlled setting. Instead, foundational hyperparameters were inherited directly from standard Stable Diffusion v1.5 training scripts and the respective baselines' default configurations, ensuring all methods were evaluated in their native optimization environments. All models underwent an identical base training phase and were subsequently exposed to the same few-shot adaptation protocol.

During the novel phase fine-tuning, all methods were subjected to identical global optimization constraints, specifically utilizing 100 fine-tuning steps and full-batch updates. Preliminary empirical investigations indicated that fewer steps generally result in underfitting, whereas exceeding 100 steps yields diminishing returns or slight degradation. Thus, 100 steps serve as an optimal, neutral checkpoint that balances concept adaptation with representation stability. Furthermore, under the extreme 5-shot constraint, utilizing a full-batch update is crucial to minimize gradient variance. This provides a stable optimization direction and prevents noisy updates that would otherwise severely destabilize the training process for all models.

Under these strictly consistent settings, the only method-specific hyperparameter adjustments, such as the $100\times$ learning rate multiplier for gating parameters $\eta$ and $\gamma$, the regularization coefficient $\lambda=0.1$, and the sensitivity threshold $\mu=0.95$, are exclusively tied to our newly introduced disentanglement modules. Since the baseline methods lack these specialized architectural components, they naturally do not require these parameters. The performance gap observed under this unified protocol ultimately reflects the vulnerability of inherently coupled baseline architectures under extreme few-shot pressure, further validating the robustness of our proposed framework. A comprehensive summary of the exact experimental configurations is provided in Table \ref{tab:hyperparameters}.

\begin{table}[h]
  \caption{Detailed hyperparameter settings for base training and few-shot fine-tuning across all evaluated methods.}
  \label{tab:hyperparameters}
  \begin{center}
    \begin{small}
      \begin{sc}
        \setlength\tabcolsep{20pt} 
        \renewcommand{\arraystretch}{1} 
        \resizebox{1\linewidth}{!}{
        \begin{tabular}{llcl}
          \toprule
          Hyperparameter Category & Hyperparameter Name & Value & Applied To \\
          \midrule
          \textbf{Global Optimization} & & & \\
          \quad (Shared) & Optimizer & AdamW & All methods \\
          \quad & Base Learning Rate & 1 $\times$ 10$^{-4}$ & All methods \\
          \quad & Resolution & 512 $\times$ 512 & All methods \\
          \quad (Base Phase Only) & Total Epochs & 100 & All methods \\
          \quad & Effective Batch Size & 320 & All methods \\
          \quad (Novel Phase Only) & Fine-tuning Steps & 100 & All methods \\
          \quad & Effective Batch Size & Full Novel Set & All methods \\
          \midrule
          \textbf{Proposed Modules} & & & \\
          \quad & Gating LR Multiplier ($\eta$, $\gamma$) & 100 $\times$ & Ours Only \\
          \quad & Regularization Coef ($\lambda$) & 0.1 & Ours Only \\
          \quad & Sensitivity Threshold ($\mu$) & 0.95 & Ours Only \\
          \quad & Primitives Iterations ($N_\text{iter}$) & 50 & Ours Only \\
          \quad & Number of Primitives ($s$) & 128 & Ours Only \\
          \midrule
          \textbf{Inference} & & & \\
          \quad & Scheduler & EulerDiscrete & All methods \\
          \quad & Denoising Steps & 50 & All methods \\
          \quad & CFG Scale & 7.5 & All methods \\
          \bottomrule
        \end{tabular}
        }
      \end{sc}
    \end{small}
  \end{center}
  \vskip -0.1in
\end{table}

\section{Computational Efficiency Analysis}

Table~\ref{tab:appcomp1} presents a comprehensive comparison of model complexity and inference efficiency between our proposed method and competing methods. 
We evaluate the models based on the number of trainable parameters, inference latency, and peak GPU memory usage. 
To ensure a fair comparison, all efficiency metrics were measured on a single NVIDIA GeForce RTX 4090 GPU. 
Specifically, we report the average inference latency (measured in seconds per image) for continuously generating 100 images at a resolution of 512 \texttimes~512, alongside the peak GPU memory usage observed during this process.

As shown in the table, our method achieves a superior trade-off between computational efficiency and generation quality. 
In terms of model size, our approach utilizes 182.65M trainable parameters, which is significantly more lightweight than CC-Diff (332.09M) and comparable to CC-Diff++ (168.68M). 

Most notably, our method demonstrates exceptional inference speed, requiring only 6.17 seconds per image.
This represents a significant acceleration compared to other state-of-the-art methods, which range from 8.31 to 8.47 seconds per image. 
Although our approach incurs a marginal increase in peak GPU memory usage (10.24 GB), this cost is justified by the substantial improvements in layout controllability and image fidelity, as evidenced by the state-of-the-art mAP and FID.

\begin{table}[h]
  \caption{Comparison of trainable parameters, inference latency, and peak GPU memory usage, along with performance on DIOR.}
  \label{tab:appcomp1}
  \begin{center}
    \begin{small}
      \begin{sc}
        \setlength\tabcolsep{10pt} 
        \renewcommand{\arraystretch}{1} 
        \resizebox{1\linewidth}{!}{
        \begin{tabular}{lcccc}
          \toprule
          Metric & MIGC \citeyearpar{migc} & CC-Diff \citeyearpar{ccdiff}  & CC-Diff++ \citeyearpar{ccdiffpp} & \textbf{Ours} \\
          \midrule
          Trainable Parameters (M) & 57.50 & 332.09 & 168.68 & 182.65 \\
          Inference Latency (Seconds Per Image) & 8.39 & 8.47 & 8.31 & 6.17 \\
          Peak GPU Memory Usage (GB) & 7.47 & 9.89 & 9.50 & 10.24 \\
          \midrule
          mAP (\%) $\uparrow$ & 22.75$\pm$ 0.77 & 24.91$\pm$ 0.77 & 24.63$\pm$ 0.84 & 26.06$\pm$ 0.89 \\
          AP\textsubscript{50} (\%) $\uparrow$ & 51.73$\pm$ 1.35 & 55.27$\pm$ 1.40 & 54.60$\pm$ 1.44 & 57.22$\pm$ 1.30 \\
          AP\textsubscript{75} (\%) $\uparrow$ & 16.14$\pm$ 0.98 & 19.21$\pm$ 0.97 & 18.71$\pm$ 1.08 & 20.46$\pm$ 1.15 \\
          FID\textsubscript{Boot} $\downarrow$ & 89.20$\pm$ 0.93 & 82.51$\pm$ 0.86 & 82.62$\pm$ 0.95 & 74.34$\pm$ 0.95 \\
          \bottomrule
        \end{tabular}
        }
      \end{sc}
    \end{small}
  \end{center}
  \vskip -0.1in
\end{table}

\section{More Qualitative Comparisons}

\begin{figure*}[!t]
  \vskip 0.2in
  \begin{center}
    \centerline{\includegraphics[width=0.72\linewidth]{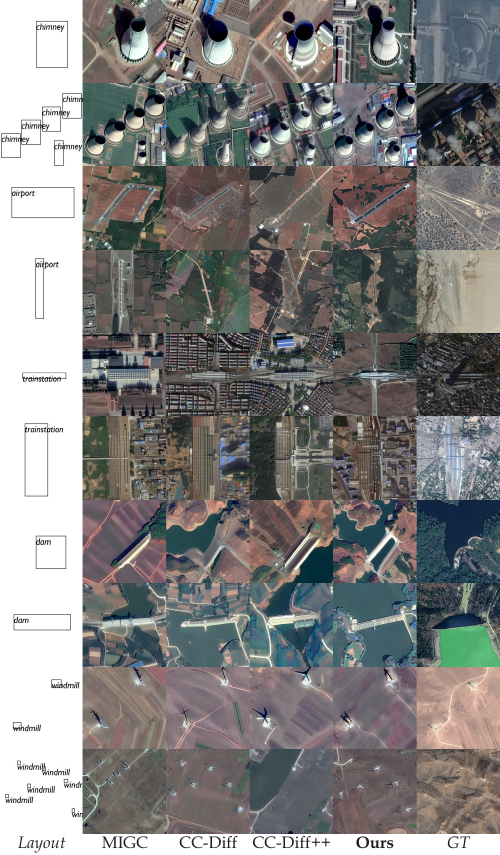}}
    \caption{
More Qualitative comparisons under the 5-shot setting on aerial domains.
    }
    \label{fig:viz-a1-dior}
  \end{center}
\end{figure*}

\begin{figure*}[!t]
  \vskip 0.2in
  \begin{center}
    \centerline{\includegraphics[width=0.72\linewidth]{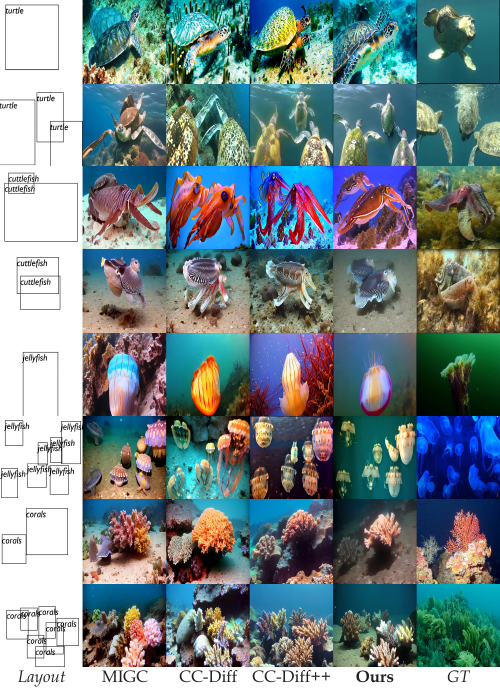}}
    \caption{
More Qualitative comparisons under the 5-shot setting on underwater domains.
    }
    \label{fig:viz-a1-ruod}
  \end{center}
\end{figure*}

\begin{figure*}[!t]
  \vskip 0.2in
  \begin{center}
    \centerline{\includegraphics[width=0.72\linewidth]{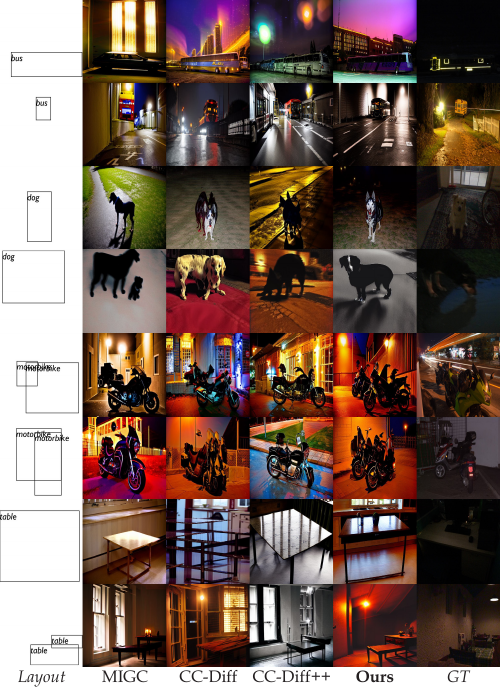}}
    \caption{
More Qualitative comparisons under the 5-shot setting on extreme dark domains.
    }
    \label{fig:viz-a1-exdark}
  \end{center}
\end{figure*}

\Cref{fig:viz-a1-dior}, \cref{fig:viz-a1-ruod}, and \cref{fig:viz-a1-exdark} present more qualitative comparisons with competing methods. 
These datasets cover diverse and challenging scenarios, including aerial views and low-light conditions, which test the generalization ability of the models.
As illustrated in \Cref{fig:viz-a1-dior}, our method exhibits superior detailed synthesis compared to the baselines. We observe that our model generates chimneys with better structural integrity, airports with precise spatial alignment, and train stations with rich textural diversity. Moreover, it captures semantically plausible dams and ensures accurate object counts of windmills. These advantages are consistently observed across \Cref{fig:viz-a1-ruod} and \Cref{fig:viz-a1-exdark}.
These additional visualizations consistently align with the quantitative results reported in the main paper.

\end{document}